\documentclass[runningheads]{llncs}
\usepackage[T1]{fontenc}
\usepackage{graphicx}
\usepackage{color}

\usepackage{cite}
\usepackage{amsmath,amssymb,amsfonts}
\usepackage{algorithmic}
\usepackage{hyperref}
\usepackage{textcomp}
\usepackage{xcolor}
\usepackage{tabularx}
\usepackage{makecell}
\usepackage{comment}
\usepackage{float}
\usepackage{graphicx}
\usepackage{multirow,array}
\usepackage{caption}
\usepackage{subcaption}
\usepackage{paralist}
\usepackage{listings}
\usepackage{zed-csp}
\usepackage{fancyhdr}
\usepackage{csquotes}
\usepackage{color}
\usepackage{upgreek} 
\usepackage{bm}
\usepackage{setspace}
\usepackage{booktabs}
\usepackage{mathtools}
\usepackage{longtable}
\begin{document}
\title{Not So Robust After All: Evaluating the Robustness of Deep Neural Networks to Unseen Adversarial Attacks}
\titlerunning{Not So Robust After All}
%
%
\author{Roman Garaev\inst{1} \and
Bader Rasheed\inst{1} \and
Adil Khan\inst{1,2} 
}
\authorrunning{Garaev et al.}
%

\institute{Innopolis University, Innopolis, Russia \email{\{o.garaev, b.rasheed\}@innopolis.university}\and
University of Hull, Hull, UK \email{a.m.khan@hull.ac.uk}}
%
\maketitle              
\begin{abstract}
Deep neural networks (DNNs) have gained prominence in various applications, such as classification, recognition, and prediction, prompting increased scrutiny of their properties. A fundamental attribute of traditional DNNs is their vulnerability to modifications in input data, which has resulted in the investigation of adversarial attacks. These attacks manipulate the data in order to mislead a DNN.
This study aims to challenge the efficacy and generalization of contemporary defense mechanisms against adversarial attacks. Specifically, we explore the hypothesis proposed by Ilyas et. al \cite{11}, which posits that DNN image features can be either robust or non-robust, with adversarial attacks targeting the latter. This hypothesis suggests that training a DNN on a dataset consisting solely of robust features should produce a model resistant to adversarial attacks. However, our experiments demonstrate that this is not universally true. To gain further insights into our findings, we analyze the impact of adversarial attack norms on DNN representations, focusing on samples subjected to \(L_2\) and \(L_{\infty}\) norm attacks. Further, we employ canonical correlation analysis, visualize the representations, and calculate the mean distance between these representations and various DNN decision boundaries. Our results reveal a significant difference between \(L_2\) and \(L_{\infty}\) norms, which could provide insights into the potential dangers posed by \(L_{\infty}\) norm attacks, previously underestimated by the research community.
 
\keywords{machine learning \and deep learning \and adversarial attacks}
\end{abstract}

\section{Introduction}
The growth of computing power and data availability has led to the development of more efficient pattern recognition techniques, such as machine learning and its subclasses, neural networks and deep learning. When a sample accurately represents a data population and is of sufficient size, machine learning methods can yield impressive results, even on unseen data, making them suitable for tasks such as classification and prediction tasks \cite{13, 27}. Although these methods offer significant advantages, they also come with limitations, such as the sensitivity of deep neural networks (DNNs) to data quality and source, and overfitting when trained on insufficient amounts of data. Researchers continue to develop new approaches and techniques to address these challenges and improve the overall effectiveness of pattern recognition \cite{17, 18, 19, 20}. Despite these attempts, variations in input samples from different domains or insufficient training data can still significantly impact DNNs’ performance. As DNNs are increasingly being employed in critical applications such as medicine and transportation, enhancing their robustness is essential due to the potentially severe consequences of their unpredictability.

One approach to examine DNN robustness to diverse inputs is through the use of adversarial attacks for analysis or training. Adversarial attacks aim to generate the smallest adversarial perturbation - a change in input that results in a misclassification by the model.

This study investigates the abilities and drawbacks of modern defense techniques against adversarial attacks, such as adversarial and robust training. The latter one refers to the hypothesis proposed by Ilyas et al. \cite{11}, which suggests that adversarial attacks exploit non-robust features inherent to the dataset rather than the objects in the images. According to this hypothesis, removing these features from the dataset and training a model on the modified data should render adversarial attacks ineffective. We replicate the experiments from the original paper, training a model on robust features and testing it on unseen attacks. Our tests reveal that the models trained on robust features are generally not resistant to \(L_{\infty}\) norm perturbations.

To explain this behavior, we compare the representations of adversarial examples using canonical correlation analysis (CCA) and discover that \textbf{\(L_{\infty}\) norm attacks cause the most dispersion in the latent representation}. We visualize neural network representations under  \(L_2\) and \(L_{\infty}\) norm attacks to illustrate the impact of norms on the distributions of representations. Additionally, we tested the corresponding adversarial trained models used in robust training, which provided insight into the relationship between robust and adversarial training, suggesting that robust training is a specific case of adversarial training.

These insights might be used by the researchers for development of more robust neural networks by adversarial training: while \(L_2\) and \(L_{\infty}\) norm perturbations look similar, the \(L_{\infty}\) norm ones are much harder to resist. As far as we know, no one paid such attention to the differences between the attack norms before.

The structure of this paper is as follows. Section \ref{sec1} presents a brief overview of various attacks on image classifiers and discusses potential defense strategies that we use in our experiments. Section \ref{sec2} explores current hypotheses regarding the nature of adversarial examples. Section \ref{sec3} details the experimental methodology and the results. The conclusion of our study is presented in Section \ref{sec5}.

\section{Related works}\label{sec1}
We focus on the attacks on image classifiers as they are the most widespread and mature, however, adversarial attacks are not limited by the type of input or task. The reader may find the examples of attacks in other domains, such as malicious URLs classification \cite{57}, communication systems \cite{28}, time series classification\cite{33}, malware detection in PDF files \cite{22}, etc. 

In this research, we suppose that an adversary has complete information about the neural network, including weights, gradients, and other internal details (white-box scenario). We also assume that in most of the cases the adversary's goal is to simply cause the classifier to produce an incorrect output, without specifying a particular target class (untargeted attack scenario). Other possible scenarios can be found, for example, in \cite{59}. 
\subsubsection{Fast gradient sign method (FGSM)}

\par FGSM  was proposed by Goodfellow et al. in \cite{15}. Adversarial example \(\tilde{x}\) for image \(x\) is calculated as:
\begin{equation}\label{FGSM}
\tilde{x} = x + \epsilon Sign(\nabla_x J(x,\theta, y))
\end{equation}
where \(\epsilon\) is the perturbation,  \(J\) is the cost function for neural network with weights \(\theta\), calculated for the input image \(x\) with true classification label \(y\). 

\subsubsection{Projected gradient descent (PGD)} 

\par PGD attack, introduced by Madry et al. in \cite{14}, is an iterative variant of FGSM, carrying out the similar operation as (\ref{FGSM}) with projection on the \(\epsilon\)-ball:
\begin{equation}\label{PGD}
\tilde{x}^t = \pi_{x+S}(\tilde{x}^{t-1} + \epsilon Sign(\nabla_x J(x,\theta, y)))
\end{equation}
where \(\pi\) is the projection of an adversarial example on the set of possible perturbations \(S\), and \(t\) is the number of step in iteration. 

\subsubsection{Carlini-Wagner (C-W) attack} 

\par Carlini and Wagner \cite{6} introduced a targeted attack which solves the adversarial optimization problem without using \(Sign(\nabla)\). The authors derived a new optimization task for adversarial attack which could be solved by a regular optimizer (for example, SGD) and does not constrain \(\tilde{x} \in {[0, 1]}^m\).

\subsubsection{DeepFool} \label{DeepFool}

\par Moosavi-Dezfooli et al. \cite{43} provided a simple iterative algorithm to perturb images to the closest wrong class. In other words, DeepFool is equal to the orthogonal projection onto the classifier's decision boundary. This property allows to use the attack to test the robustness of a models:
\begin{equation}
\hat{\rho}_{adv}(f) = \frac{1}{|D|}\sum_{x \in D} \frac{\|\hat{r}\|_2}{\|x\|_2}
\end{equation}
 where \(\hat{\rho}_{adv}\) - is the average robustness, \(f\) - is the classifier, \(\hat{r}\) - is the successful perturbation from DeepFool, \(D\) - is the dataset. DeepFool can be used to calculate the distance from a data point to the closest point on the decision boundary \cite{44}. 

\subsection{Adversarial training as a defense method}
One of the most popular approaches to defending the neural networks against attacks is called \textit{adversarial training}. It proposes to "include" possible adversarial examples into the training data sets to prepare a model for the attacks. The following min-max optimization problem \cite{14} should be solved to get the adversarial trained model:
\begin{equation} \label{adv_train}
\underset{{\theta}}{\arg\min} \sum_{(x_i, y_i) \in D}\underset{\delta \in S}{\max}L(f_{\theta}(x_i + \delta), y_i)
\end{equation}
\newline where, \(\theta\) - are the weights of neural network,  \(D\) - the training data set, \(S\) - the space of possible perturbations, \(S = \{\delta: \|\delta\|_p < \epsilon\}\) for a given radius \(\epsilon\). 

Adversarial training was introduced by Goodfellow et al. in \cite{15}. Madry et al. \cite{14} proposed using a PGD attack during the training procedure and presented better robustness against adversarial attacks. However, Wong et al. \cite{12} achieved about the same accuracy against adversarial attacks, using simple one-step FGSM.

It is important to note that the \textit{provable} defense (i.e. "certified robust") against \textit{any} small-\(\epsilon\) attack has already been studied, for example, by Wong and Kolter \cite{52} and Wong et al. \cite{53}. However, the experiments in these works were conducted with relatively small perturbations: in \cite{53} the maximum radius of \(L_{\infty}\) perturbation is \(\frac{2}{255}\), while the same norm in adversarial training package \cite{37} is \(\frac{8}{255}\). Thus, we do not include certified robust methods in our research.

\section{Hypothesis about the cause of adversarial attacks}\label{sec2}
The exact reason why neural networks are susceptible to small changes in input data remains unclear. Goodfellow et al. \cite{15} argued that adversarial examples result from models being overly linear rather than nonlinear. However, another perspective considers poor generalization as the source of attacks. Ilyas et al. \cite{11} hypothesized that neural networks' vulnerability to adversarial attacks arises from their data representation. Classifiers aim to extract useful features from data to minimize a cost function. Ideally, these features should be related to object classification (\textit{robust features}), but neural networks may utilize unexpected properties specific to a particular dataset (\textit{non-robust features}), which can be manipulated by an adversary. If a classifier can be trained on a dataset containing only robust features, it should be resistant to adversarial attacks. We refer to this process as \textit{robust training}.

We decided to challenge this hypothesis for several reasons. First, it is not fully proven, except for the toy example in \cite{11} and experiments on robust and non-robust dataset creation. Second, subsequent works like \cite{55} and \cite{56} consider this hypothesis, while it might not be entirely accurate. For example, Zhang et al. \cite{56} proposed the similar experiment, referring to \cite{11}, and developed it onto universal perturbation.  Although the results of \cite{11} was discussed in \cite{36}, the accuracy of robust trained model was tested nowhere but in the original work, and we would like to fill this gap.

By testing this hypothesis, we discover certain properties of adversarial attacks and further explore them through our experiments. We found not only cases where robust features can be corrupted, but also that \(L_{\infty}\)-norm attacks are more dangerous than those with \(L_2\)-norm. We hypothesize that even if perturbations of these two norms are both imperceptible to humans, \(L_{\infty}\)-norm attacks have a greater impact on model representations.

\section{Experiments} \label{sec3}
\par In this work, we establish the following Research Aims (RAs):
\begin{itemize}
\item \textbf{(RA 1)} Our research aims to explore the generalization of robust and adversarial training methods. Specifically, we investigate the classification accuracy of robust and adversarial trained models when subjected to various unseen attacks. Additionally, we aim to compare the behavior of these models under identical attack conditions and develop a methodology to measure such differences.

\item \textbf{(RA 2)} Our study also seeks to identify potential shortcomings of robust trained models. We aim to determine the proximity of adversarial and benign samples in the latent space of a robust trained model, i.e. measure the distance between the representations. The degree of similarity between representations of samples with small \(L_{\infty}\) or \(L_2\) norms would indicate the stability of a model.

\item \textbf{(RA 3)} Addressing the previous RA, we aim to investigate the impact of robust and adversarial training methods on the decision boundary of a model. We seek to determine how the mean distance between samples and decision boundaries varies for different models when exposed to adversarial attacks.
\end{itemize}

To address the research aims, we present a series of experiments. In the first experiment, we replicate robust training from \cite{11}, employing a broader testing setup that included various attack norms, data sets, and model architectures. To compare the performance of adversarial and robustly trained models, we conduct the Kolmogorov-Smirnov test. Next, we analyze the representations of neural networks from the first experiment by singular value canonical correlation analysis (SVCCA) and perform principal component analysis (PCA) for their visualization. Finally, we assess the mean distance from samples to the decision boundaries of different models using the DeepFool attack. In the rest of the section, we describe in details the experiment setups and the achieved results.

\subsection{The broad testing of robust and adversarial training (RA 1)}
\par Here and later, we refer to training based on empirical risk minimization as "regular" since it does not involve any adversarial attack and is traditionally used to train neural networks. 

The motivation for these experiments stems from the work of Tramer et al. \cite{4}, which demonstrated that even adversarially trained models could be compromised by unseen attacks. The core premise of this experiment involves taking an adversarially trained deep model, trained to defend against a specific attack on a given dataset, and utilizing it to develop the robust version of this model, and then evaluate the generalisation perofmance of both models against unseen attacks. This process is undertaken as follows:

\begin{enumerate}
\item We select every image within the dataset and use the process outlined in \cite{11} to extract an image that only possesses robust features. The adversarially trained model aids in this extraction. Initially, we generate a random noise image for each target image. Then, we iteratively compute the representations of both the random and target images as interpreted by the pre-trained model. At each iteration, we slightly adjust the random image to minimize the distance between the vectors of the two images. After a set number of steps, this method produces a modified dataset.
\item We then proceed to train the regular version of our selected adversarially trained model on this modified dataset, ultimately resulting in a robustly trained model.
\item Subsequently, we test the generalisation capacity of both the adversarial and robust models against attacks which were not considered during the adversarial training phase, and hence, not used in the formation of the robust model.
\end{enumerate}

This entire experiment is practically implemented for two distinct models: ResNet50s and InceptionV3. For ResNet50, we create both the adversarially trained and robust versions, aiming them at the CIFAR-10 \cite{46} dataset and defending against the PGD attack with \(L_2\) and \(L_{\infty}\) norms. We then test their performance against FGSM (\(L_1\), \(L_2\), \(L_{\infty}\) norms), PGD (\(L_1\), \(L_2\), \(L_{\infty}\) norms), C-W (\(L_2\) norm), and DeepFool (\(L_2\) norm) attacks. The performance results of the \(L_2\)-trained model are outlined in Table \ref{tab:attacks}, and those of the \(L_{\infty}\) trained model are presented in Table \ref{tab:attacks_inf}.

The adversarially trained and robust versions of InceptionV3, on the other hand, were trained only against the PGD attack with the \(L_{\infty}\) norm for CIFAR-10 dataset. The testing parameters for these models were akin to those used with ResNet50, and the results are displayed in Table \ref{tab:inception_2}. All tables highlight the corresponding values of \(\epsilon\)-s and the number of steps for iterative attacks used in each case. Note that \(\epsilon\) for the attacks differs from the similar ones for ResNet50-s because the model has the bigger input shape (224x224 vs 32x32, respectively).

\begin{table}[h]
    \centering
    \begin{tabular}{p{1.8cm} | p{1.2cm} | p{1.2cm} |  p{1.2cm} | p{1.2cm} |  p{1.2cm}}
         Attack & Norm & Epsilon & Steps & Robust acc. & Adv. acc.\\ \hline
         No attack & -        & -        & -  & 0.813 & 0.91 \\ 
         FGSM             & \(L_1\)    & 0.5    & 1 & 0.81   & 0.91 \\
         FGSM             & \(L_2\)    & 0.25  & 1 & 0.59   & 0.87 \\
         FGSM             & \(L_{\infty}\)  & 0.25   & 1 & 0.1    & 0.13 \\
         FGSM             & \(L_{\infty}\)  & 0.025 & 1 & 0.22  & 0.62 \\ \hline
         PGD & \(L_1\)     & 0.5   & 100    & 0.81  & 0.91\\
         PGD & \(L_2\)     & 0.25  & 1000 & 0.483 &  0.82\\
         PGD & \(L_2\)     & 0.5   & 100    & 0.202 & 0.75 \\
         PGD & \(L_{\infty}\) & 0.025 & 5      & 0.168 & 0.54\\  
         PGD & \(L_{\infty}\) & 0.25  & 5       &  0.08 & 0.06\\  \hline
         C-W & \(L_2\)     & 0.25 & 10 & 0.219  & 0.86\\
         DeepFool & \(L_{2}\) & 0.25 & - & 0.124 & 0.127\\ \hline
    \end{tabular}
    \caption{Robust ResNet50 trained on \(L_{2}\) data set and related adversarial model. Epsilon column stands for budget of perturbation, norm - the way of measurement of this budget, steps - maximum iterations for adversarial example creation,  Robust acc. - accuracy of robust trained model, Adv. acc. - accuracy of the corresponding adversarial trained model}
    \label{tab:attacks}
\end{table}

\begin{table}[h]
    \centering
    \begin{tabular}{p{1.8cm} | p{1.2cm} | p{1.2cm} |  p{1.2cm} | p{1.2cm} |  p{1.2cm}}
         Attack & Norm & Epsilon & Steps & Robust acc. & Adv. acc.\\ \hline
         No attack & -        & -        & -  & 0.73 &  0.87\\
         FGSM             & \(L_1\)    & 0.5    & 1 &  0.73  & 0.87  \\
         FGSM             & \(L_2\)    & 0.25  & 1 &  0.504  &  0.826\\
         FGSM             & \(L_{\infty}\) & 0.25   & 1 &  0.07 &  0.19\\
         FGSM             & \(L_{\infty}\) & 0.025 & 1 &  0.2 &  0.724\\ \hline
         PGD & \(L_1\)     & 0.5     & 100   & 0.731  & 0.87\\
         PGD & \(L_2\)     & 0.25   & 1000 & 0.414  & 0.813\\
         PGD & \(L_2\)     & 0.5     & 100   & 0.195  &  0.663\\
         PGD & \(L_{\infty}\)     & 0.025 & 5       & 0.155  & 0.683\\  
         PGD & \(L_{\infty}\)     & 0.25   & 5       & 0.11    & 0.052\\  \hline         
         C-W & \(L_2\)     & 0.25 & 10 & 0.51 &  0.81\\        
         DeepFool & \(L_{2}\) & 0.25 & - & 0.13 & 0.111\\ \hline
    \end{tabular}
    \caption{Robust ResNet50 trained on \(L_{\infty}\) data set and related adversarial model}
    \label{tab:attacks_inf}
\end{table}

\begin{table}[h]
    \centering
    \begin{tabular}{p{1.8cm} | p{1.2cm} | p{1.2cm} |  p{1.2cm} | p{1.2cm} |  p{1.2cm}}
         Attack & Norm & Epsilon & Steps & Robust acc. & Adv. acc.\\ \hline
         No attack        & -          & -     & - & 0.89  & 0.94 \\
         FGSM             & \(L_2\)    & 0.25  & 1 & 0.49  & 0.83 \\
         FGSM             & \(L_{\infty}\) & 0.25   & 1 &  0.10 &  0.415\\
         FGSM             & \(L_{\infty}\) & 0.025 & 1 & 0.63   & 0.79 \\ \hline
         PGD & \(L_2\)     & 100 & 5       &  0.87  & 0.912\\
         PGD & \(L_2\)     & 45   & 100   & 0.88   & 0.903 \\
         PGD & \(L_2\)     & 0.5  & 100   &  0.89  & 0.94 \\
         PGD & \(L_{\infty}\)    & 0.5  & 5       &  0.06  & 0.378  \\  
         PGD & \(L_{\infty}\)    & 1.0  & 5       &   0.04  & 0.281 \\  \hline         
         C-W & \(L_2\)  & 0.25 & 10 & 0.51 & 0.85\\        
         DeepFool & \(L_{2}\) & 0.25 & - &  0.48  & 0.784\\ \hline
    \end{tabular}
    \caption{Robust InceptionV3 trained on \(L_{\infty}\) data set}
    \label{tab:inception_2}
\end{table}

On examination, it's apparent that both tested models demonstrate reasonable stability against some variations of PGD and FGSM attacks. However, all attacks with an \(L_{\infty}\) norm significantly undermined the accuracy of the models. Hence, the model doesn't ensure generalisation against all attack types, primarily as a simple increase in norm or perturbation shows a drastic impact. Despite this, the models did demonstrate resistance to certain unseen attacks, particularly those with an \(L_1\) norm, exhibiting only a minor drop in accuracy. This resistance to \(L_1\) norm attacks is crucial for these models in terms of generalisation, given they weren't exposed to such an attack during training.

The accuracy of adversarially and robustly trained models might vary, but a similar ratio between them implies similar behavior. We evaluate this accuracy employing the Kolmogorov-Smirnov test for goodness. The null hypothesis \(H_0\) at a significance level of 0.05 asserts that the samples, i.e., \textit{accuracy of robust and adversarial trained models under different attacks, are from the same distribution}. The test for accuracy presented in Table \ref{tab:attacks} illustrates that, with a \(p-value of 0.42\), we \textbf{cannot reject the \(H_0\) hypothesis}. Therefore, robust models may be considered as a specific case of adversarial training, with its strengths and weaknesses.

In a bid to further extend this experiment, we opt to manage the entire adversarial training pipeline ourselves, thereby performing the process from scratch. Owing to computational constraints, we select the relatively more manageable ResNet18 architecture. The models are trained on the PGD attack with 5 iterations and \(L_2\) and \(L_{\infty}\) norms. For datasets, we utilize CIFAR-10 (the results are displayed in Table \ref{tab:resnet18_1}) and CINIC-10 \cite{47} (refer to Table \ref{tab:resnet18_2}), and for test attacks, we use PGD and FGSM. Consistent with the previous pattern, attacks with the L-inf norm result in a more significant drop in accuracy than similar ones with the L2 norm.

\begin{table*}[h]
    \centering
    \begin{tabular}{ p{3cm} | p{2.8cm} | p{2.8cm} |  p{2.8cm} }
     & Regular model &  Adv.trained, \(L_{\infty}\) norm  & Adv. trained, \(L_{2}\) norm\\ \hline
  Accuracy on CINIC-10 (no attack) & 76 \% & 75 \% & 72 \% \\
  Accuracy on CIFAR-10 (no attack) & 95 \% & 94 \% & 91  \% \\ \hline
  Accuracy on CINIC-10 (PGD attack) & 7\% - \(L_{\infty}\) attack, 11\% - \(L_2\) attack & 37\% - \(L_{\infty}\) attack, 36\% - \(L_2\) attack & 40 \% - \(L_{\infty}\) attack, 43\% - \(L_2\) attack \\
  Accuracy on CIFAR-10 (PGD attack) & 7\% - \(L_{\infty}\) attack, 27\% - \(L_2\) attack & 57\% - \(L_{\infty}\) attack, 55\% - \(L_2\) attack & 61 \% - \(L_{\infty}\) attack, 64 \% - \(L_2\) attack \\ \hline
  Accuracy on CINIC-10 (FGSM attack) & 49 \% - \(L_{\infty}\) attack & 67 \% - \(L_{\infty}\) attack & 66 \% - \(L_{\infty}\) attack \\
  Accuracy on CIFAR-10 (FGSM attack) & 72 \% - \(L_{\infty}\) attack & 89 \% - \(L_{\infty}\) attack & 87 \% - \(L_{\infty}\) attack \\ \hline
    \end{tabular}
    \caption{Accuracy of ResNet18-s, trained on CIFAR-10 data set}
    \label{tab:resnet18_1}
\end{table*}

\begin{table*}[h]
    \centering
    \begin{tabular}{ p{3cm} | p{2.8cm} | p{2.8cm} |  p{2.8cm} }
     & Regular model &  Adv.trained, \(L_{\infty}\) norm  & Adv. trained, \(L_{2}\) norm \\ \hline
   Accuracy on CINIC-10 (no attack) & 86 \% & 84 \% & 80 \% \\
  Accuracy on CIFAR-10 (no attack) & 94 \% & 93 \% & 90  \% \\ \hline
  Accuracy on CINIC-10 (PGD attack) & 3\% - \(L_{\infty}\) attack, 6\% - \(L_2\) attack & 33\% - \(L_{\infty}\) attack, 30\% - \(L_2\) attack & 42 \% - \(L_{\infty}\) attack, 46\% - \(L_2\) attack \\
  Accuracy on CIFAR-10 (PGD attack) & 4\% - \(L_{\infty}\) attack, 7\% - \(L_2\) attack & 45\% - \(L_{\infty}\) attack, 41\% - \(L_2\) attack & 55 \% - \(L_{\infty}\) attack, 59 \% - \(L_2\) attack \\ \hline
  Accuracy on CINIC-10 (FGSM attack) & 50 \% - \(L_{\infty}\) attack & 75 \% - \(L_{\infty}\) attack & 73 \% - \(L_{\infty}\) attack \\
  Accuracy on CIFAR-10 (FGSM attack) & 63 \% - \(L_{\infty}\) attack & 87 \% - \(L_{\infty}\) attack & 85\% - \(L_{\infty}\) attack \\ \hline
    \end{tabular}
    \caption{Accuracy of ResNet18-s, trainied on CINIC-10 data set}
    \label{tab:resnet18_2}
\end{table*}

\subsection{Comparison of representations under adversarial attacks (RA 2)}
Canonical correlation analysis (CCA) is a method utilized to compare the representations of neural networks. Its objective is to identify linear combinations of two sets of random variables that maximize their correlation. In previous studies \cite{39}, \cite{40}, CCA has been employed to compare activations from different layers of neural networks. Raghu et al. \cite{41} proposed an extension of CCA, singular value canonical correlation analysis (SVCCA), for the analysis of neural networks. 

In this study, we employ SVCCA to compare the representations of the original and corresponding adversarial images. For each experiment, we take a batch of 128 images, compute the related adversarial examples under some attack, calculate SVCCA for the representations, and take the mean. We test three models in each experiment: regular and two adversarial ResNet50-s trained with PGD using \(L_{2}\) and \(L_{\infty}\) norms. In this experiment, a high mean correlation coefficient indicates that the representations are similar to each other and that small perturbations do not impact the model.

While the attack norms are the primary variables in these experiments, we also test two different attacks (FGSM and PGD) to eliminate the threat of validity. The means of SVCCA for the attacks are presented in Table \ref{tab:cca}, while Table \ref{tab:cca_2} shows the same for the best and worst cases from Tables \ref{tab:attacks} and \ref{tab:attacks_inf}.

\begin{table*}[h]
    \centering
    \begin{tabular}{ p{2cm} | p{5cm} | p{1.5cm} |  p{1.5cm} |  p{1.5cm}  }
    Attack & Parameters & Regular model & Adv. trained, \(L_{2}\) & Adv.trained, \(L_{\infty}\) \\ \hline
    \multirow{2}{*}{PGD} & \(L_2\), \(\epsilon=0.25\), \(steps=100\)  & 0.686 & 0.981 & 0.989 \\ 
     & \(L_{\infty}\), \(\epsilon=0.025\), \(steps=100\)  & \textbf{0.587} &\textbf{0.751} & \textbf{0.822} \\  \hline
    \multirow{2}{*}{FGSM} &  \(L_{2}\), \(\epsilon=0.25\) &  0.833 &  0.982 & 0.991 \\ 
    & \(L_{\infty}\), \(\epsilon=0.025\) & 0.618  & 0.78 & 0.837  \\ \hline
    \end{tabular}
    \caption{Mean correlation coefficient for different models}
    \label{tab:cca}
\end{table*}

\begin{table*}[h]
    \centering
    \begin{tabular}{ p{2cm} | p{4cm} | p{1.5cm} |  p{1.5cm} |  p{1.5cm}  }
    Attack & Parameters & Regular model & Adv. trained, \(L_{2}\) & Adv.trained, \(L_{\infty}\) \\ \hline
    PGD, best    & \(L_2\), \(\epsilon=0.25\), \(steps=1000\)  & 0.682 & 0.989 &  0.99 \\
    PGD, worst & \(L_{\infty}\), \(\epsilon=0.25\), \(steps=5\)  & 0.46  & 0.532 &  0.598  \\ 
    \hline \hline
    FGSM, best    &  \(L_{2}\), \(\epsilon=0.25\) & 0.833  & 0.982 & 0.991\\ 
    FGSM, worst &  \(L_{\infty}\), \(\epsilon=0.25\) & 0.422 & 0.346 & 0.411\\ 
    \hline
    \end{tabular}
    \caption{Mean correlation coefficient for the best and worst cases in Tables \ref{tab:attacks} and \ref{tab:attacks_inf}}
    \label{tab:cca_2}
\end{table*}

Overall, the results presented in tables \ref{tab:cca} and \ref{tab:cca_2} demonstrate that the impact of adversarial attacks on models is most significant when using \(L_{\infty}\)-norm perturbation. This effect is evident even in models that have been trained specifically to handle \(L_{\infty}\)-norm attacks. These findings emphasize the importance of considering the norms of adversarial attacks when evaluating the robustness of models because testing on \(L_2\)-norm may provide a false sense of security.

\subsection{Visualization of representations (RA 2)}
The visualization of representations presents a challenge due to the high dimensionality of representation vectors. Nonetheless, such visualization can be useful for analysis purposes. To address this issue, we employ Principal Component Analysis (PCA) to reduce the dimensionality of representations from 512 to 2. We limit our experimentation to ResNet18 due to computational constraints.

\begin{figure}[h]
    \centering
    \includegraphics[scale=0.2]{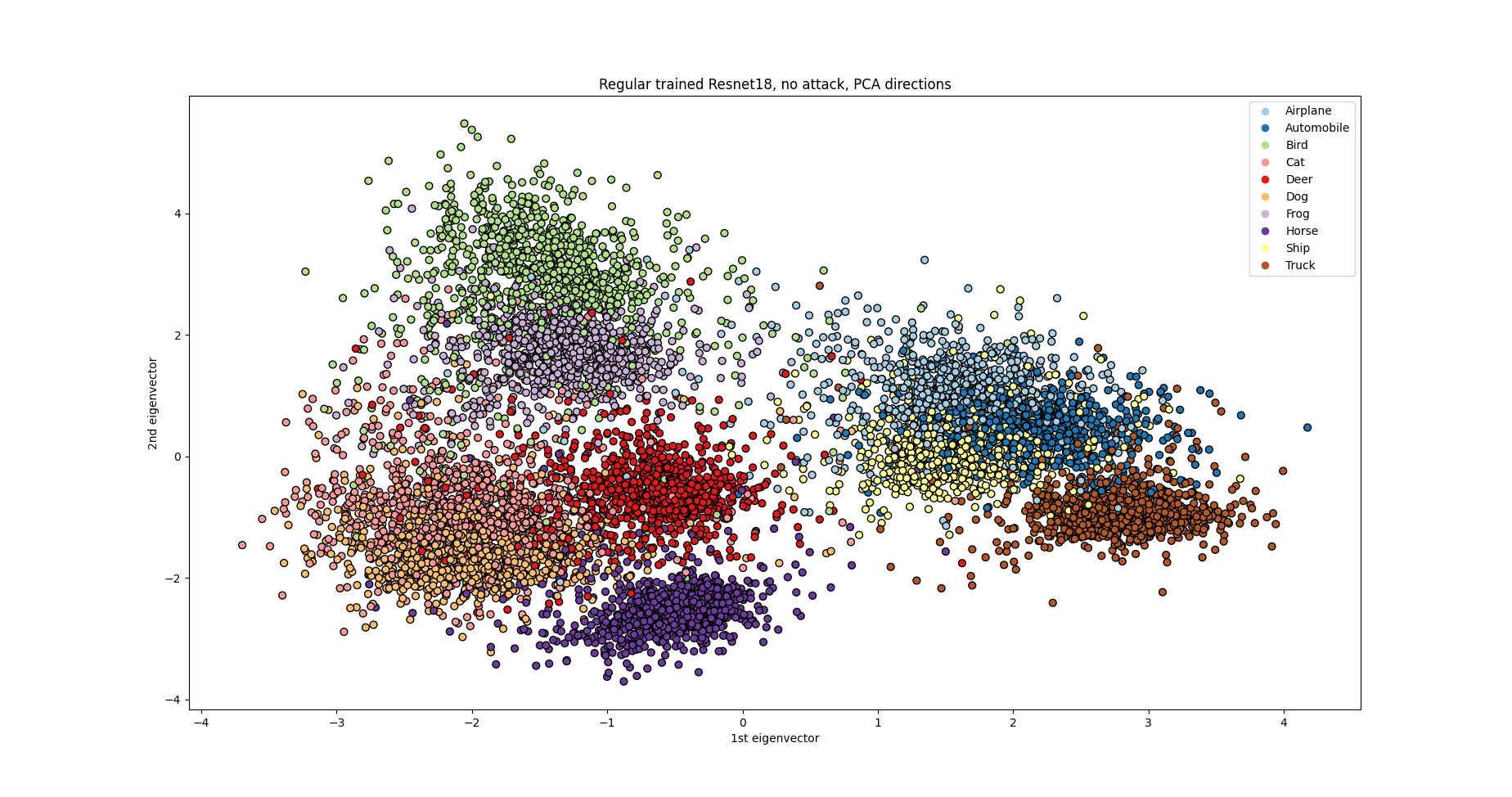}
    \caption{Representations of regular trained ResNet18 on clean data set (CIFAR-10)}
    \label{fig:label_4}
\end{figure}

\begin{figure}[h]
    \centering
    \includegraphics[scale=0.2]{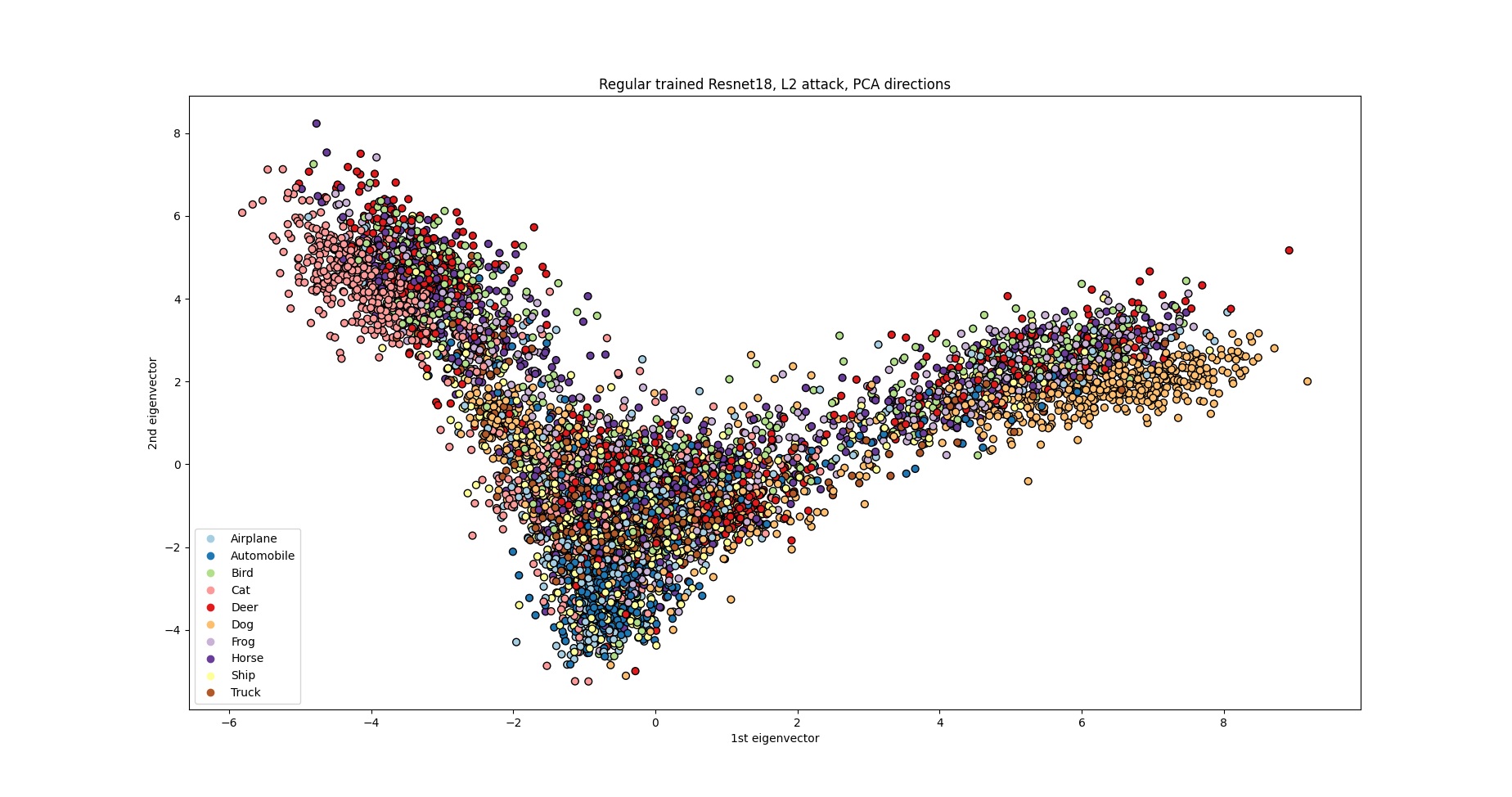}
    \caption{Representations of regular trained ResNet18 on adversarial data set (CIFAR-10), PGD with \(L_2\)-norm and 20 steps}
    \label{fig:label_5}
\end{figure}

\begin{figure}[h]
    \centering
    \includegraphics[scale=0.2]{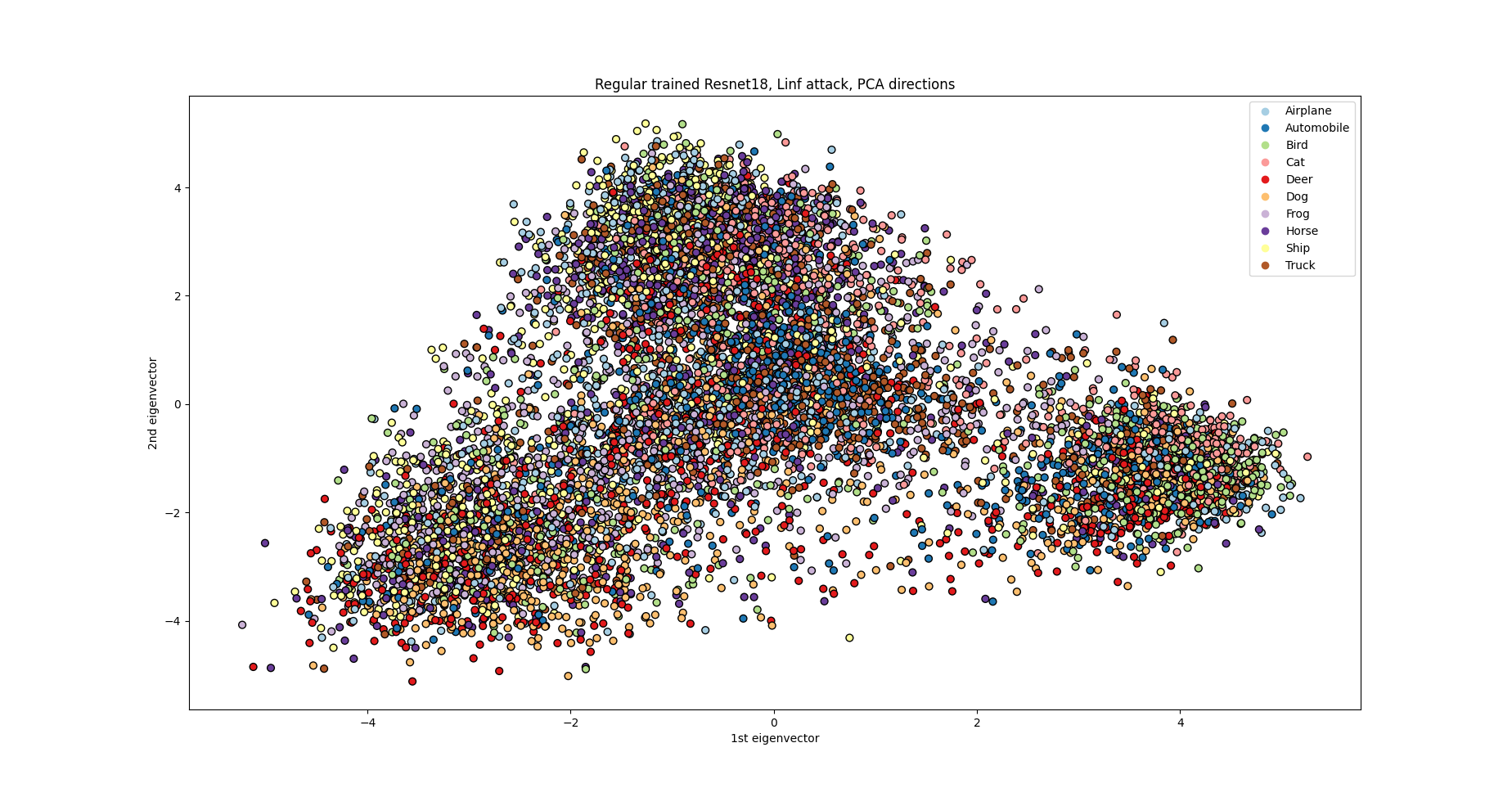}
    \caption{Representations of regular trained ResNet18 on adversarial data set (CIFAR-10), PGD with \(L_{\infty}\)-norm and 20 steps}
    \label{fig:label_6}
\end{figure}

\begin{figure}[h]
    \centering
    \includegraphics[scale=0.2]{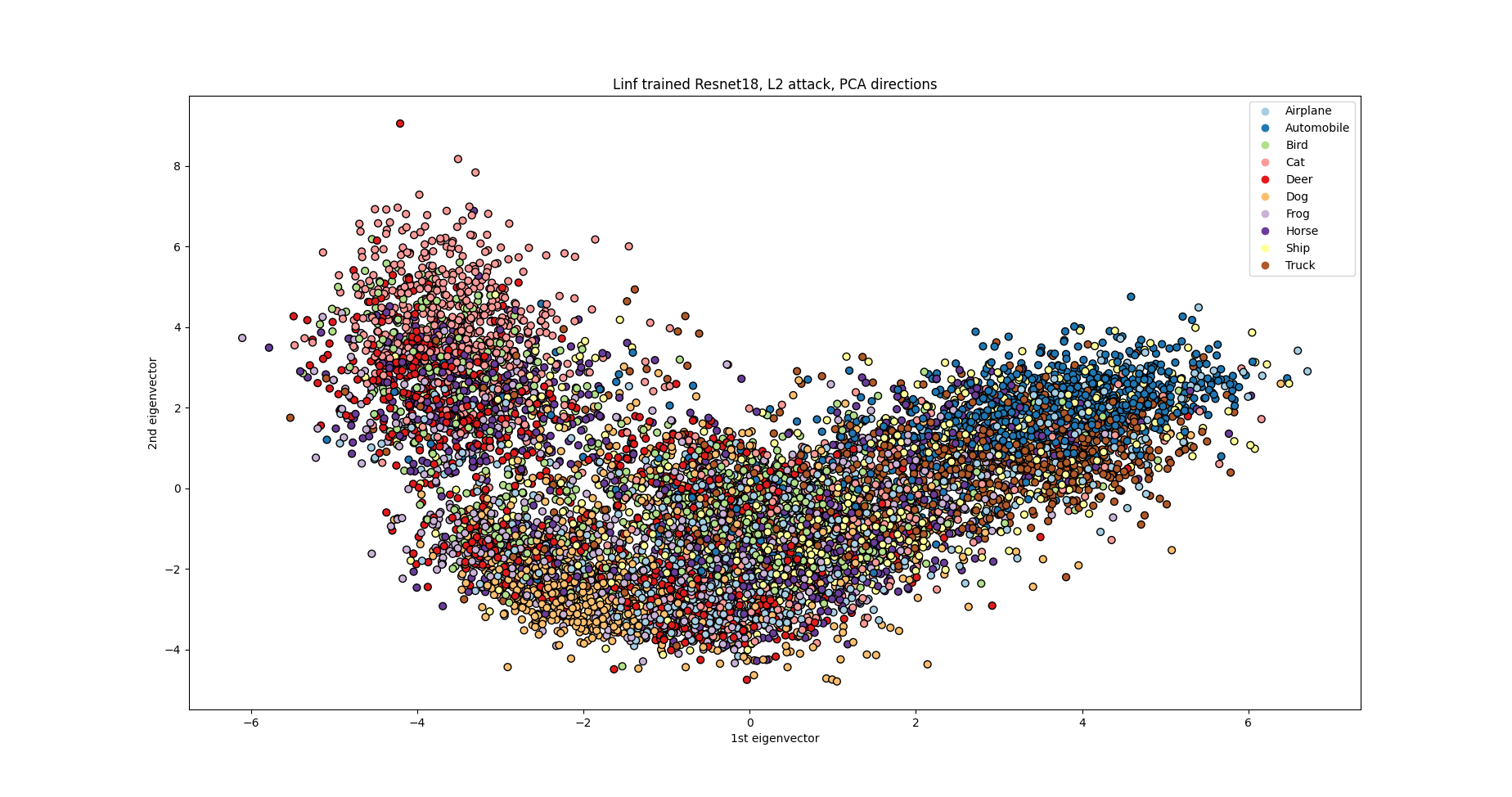}
    \caption{Representations of adversarial trained ResNet18 (PGD, \(L_{\infty}\)-norm) on adversarial data set (CIFAR-10), PGD with \(L_2\)-norm and 20 steps}
    \label{fig:label_7}
\end{figure}

\begin{figure}[h]
    \centering
    \includegraphics[scale=0.2]{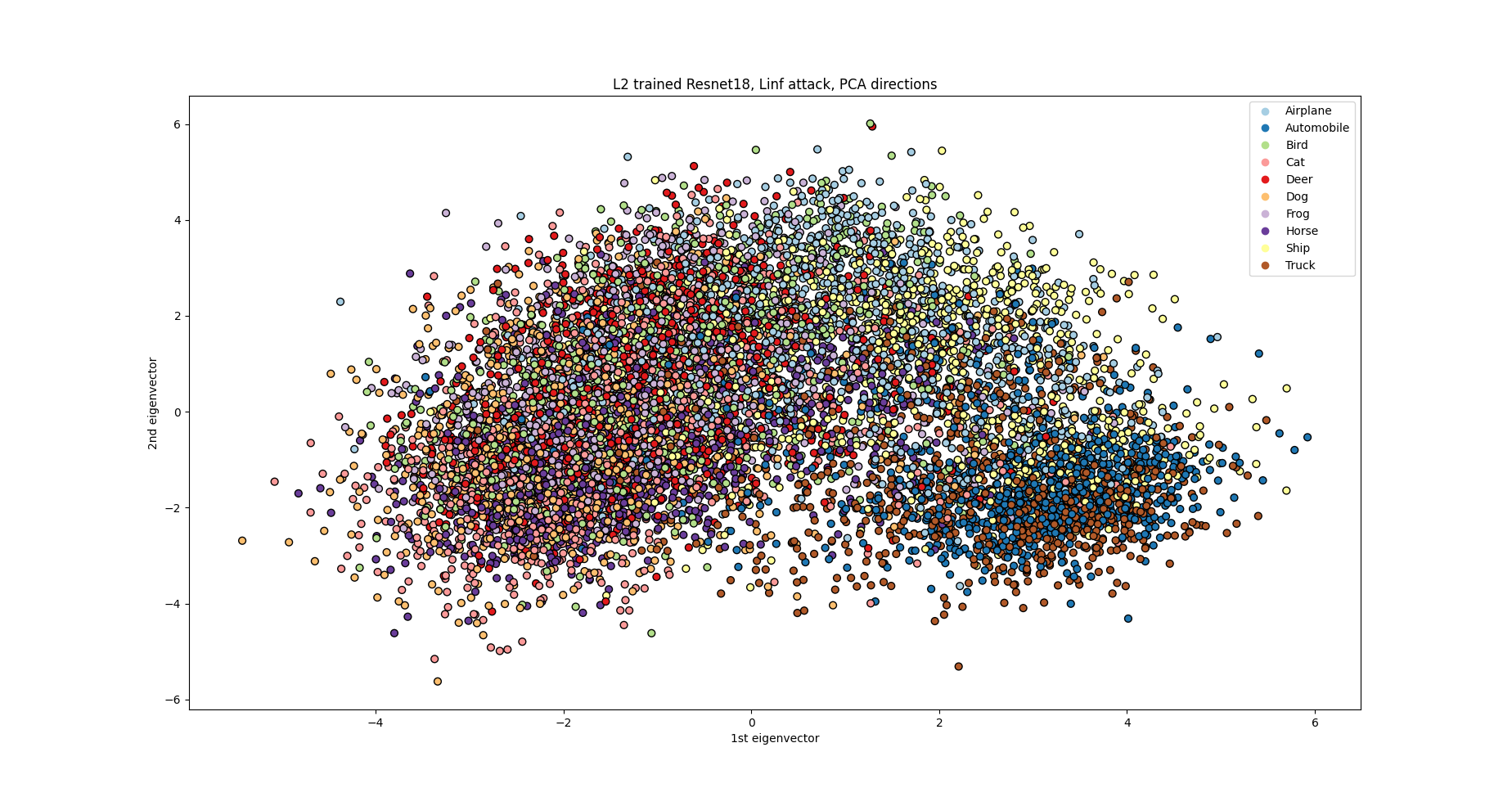}
    \caption{Representations of adversarial trained ResNet18 (PGD, \(L_2\)-norm) on adversarial data set (CIFAR-10), PGD with \(L_{\infty}\)-norm and 20 steps}
    \label{fig:label_8}
\end{figure}

We present the visualization of representations for different combinations of models and norms of PGD attacks in Figures \ref{fig:label_4} - \ref{fig:label_8}. Figure \ref{fig:label_4} depicts the representations of samples from CIFAR-10 for a regularly trained ResNet18, which serves as a baseline case. In Figures \ref{fig:label_5} and \ref{fig:label_6}, we visualize the data as adversarial samples with \(L_2\) and \(L_{\infty}\) norms of attack, respectively. Furthermore, we examine the representations of adversarial samples for adversarial trained models in Figures \ref{fig:label_7} and \ref{fig:label_8}. To challenge the models, we use alter norms of attacks from the training ones, where the model trained on PGD with \(L_2\) norm is tested on \(L_{\infty}\)-norm PGD attack (Figure \ref{fig:label_8}), and vice versa (Figure \ref{fig:label_7}). We group the representations of different classes by colors in all figures to comprehend how the representations of different classes intermingle under adversarial attacks.

The regular trained network representations are clustered according to the classes in the data set, as depicted in Figure \ref{fig:label_4}. However, when subjected to adversarial attacks, all representations become heavily mixed, resulting in a more challenging classification task due to overlapping representations of different classes. The most mixed representations were observed on the regular trained network through the \(L_{\infty}\)-norm PGD attack (Figure \ref{fig:label_6}). In contrast, adversarial training, as shown in Figures \ref{fig:label_7} and \ref{fig:label_8}, resulted in some class representations (e.g., "Automobile" or "Truck") remaining clustered while becoming closer to each other than those without attacks. The same pattern was observed for the \(L_2\)-norm attack (Figure \ref{fig:label_6}). 

Both \(L_{\infty}\) and \(L_2\) demonstrate a marked decrease in accuracy, as evidenced by the aforementioned experiments (Tables \ref{tab:resnet18_1} and \ref{tab:resnet18_2}). However, the impact of these norms on representations is not immediately apparent from a plain accuracy analysis. Indeed, the visualization of the representations distribution yielded intriguing patterns that warrant further investigation.

\subsection{Distance to decision boundary (RA 3)}
This study seeks to investigate the impact of adversarial training on the decision boundaries of models. Specifically, the mean distance between samples and the decision boundary is examined to determine how it differs for adversarially trained models. We use the idea of Mickisch et al. \cite{44} who utilized the DeepFool attack \ref{DeepFool} to measure this distance. 

The decision boundary is defined as the set of input images where two or more classes share the same maximum probability, indicating that the classifier is uncertain about the class of the image.

\begin{equation*}
\begin{aligned}
    & D =  \{ x \in R^n |  \exists k_1, k_2 = 1 ... c, k_1 \neq k_2,  \\
    & f(x)_{k_1} = f(x)_{k_2} = max (f(x))\}
\end{aligned}
\end{equation*}
where \(c\) - is the number of classes. Under this definition, the usage of the DeepFool attack looks natural because it aims to find a perturbation to the closest wrong class. The distance of the sample \(x\) to the decision boundary \(D\) can be measured as:
\begin{equation*}
    d(x) = \min \epsilon , s.t. x + \epsilon \in D
\end{equation*}

The pretrained ResNet50 model is used with 20 steps of PGD attack during training, while ResNet18 models are trained under PGD with only 5 steps. The testing dataset is CIFAR-10. The outcomes of ResNet18 and ResNet50 models with different training configurations are presented in Table \ref{tab:dist}. The table displays the mean difference between decision boundaries of models and images from CIFAR-10, calculated using both \(L_2\) and \(L_{\infty}\) distances. The "steps" column represents the mean iteration of DeepFool spent during the attack generation. The comparison is made between the models from Tables \ref{tab:resnet18_1}, \ref{tab:attacks}, \ref{tab:attacks_inf}.

\begin{table*}[]
    \centering
    \begin{tabular}{ p{6cm} | p{2.5cm} | p{2.5cm} |  p{1.5cm} |    }
    Model & Mean \(L_2\) distance & Mean \(L_{\infty}\) distance  & Steps \\ \hline
    Regular trained ResNet18 & 0.1793 & 0.0227 & 1.92 \\ \hline
    Adversarial trained ResNet18 (\(L_2\) norm) & 0.659 & 0.0824 & 1.98 \\ \hline
    Adversarial trained ResNet18 (\(L_{\infty}\)norm)  & 0.1018 & 0.0139 & 2.5 \\ \hline
    \hline \hline
    Regular trained ResNet50 & 0.17 & 0.02 & 2.58 \\ \hline
    Adversarial trained ResNet50 (\(L_2\) norm) & 1.3728 & 0.162 & 1.77 \\ \hline
    Adversarial trained ResNet50 (\(L_{\infty}\) norm)  & 1.18 & 0.303& 2.66 \\ \hline
    \end{tabular}
    \caption{Mean distance of samples to decision boundary}
    \label{tab:dist}
\end{table*}

The results indicate that the mean distance for ResNet18-s (for all types of training) to the decision boundary is relatively small, especially for \(L_{\infty}\) norm. A small distance from a sample to the decision boundary implies that it is easier to misclassify this sample because it does not require a significant perturbation. Interestingly, the distance for \(L_{\infty}\) norm adversarial trained model is actually less than training \(\epsilon\). However, testing of this model under PGD attack (Table \ref{tab:resnet18_1}) suggests that it has some robustness. Therefore, the reliability of the popular method of model testing used in this study is called into question.

\section{Conclusion} \label{sec5}
\par The experiments demonstrate that the model, trained on a "robust" data set, is still vulnerable to some attacks; thus, adversarial attacks do not compromise only non-robust features. Therefore, the robust features are not well generalized, especially on \(L_{\infty}\) norm of attack. We assume that \textit{ the small difference between clean and adversary inputs for the \(L_{\infty}\) attack leads to a huge gap in latent space between them}; SVCCA analysis of different attack representations confirms this assumption. Our visualization of neural network representation also displays the difference between \(L_2\) and \(L_{\infty} \) norms of attack. 

In light of these results, we recommend that researchers in the field of adversarial attacks and defense mechanisms pay closer attention to \(L_{\infty}\)-norm attacks to avoid a false sense of security. It is crucial to consider this norm in their tests to ensure the robustness of models against potential attacks.

%
%
%
\bibliographystyle{splncs04}
%

\bibliography{samplepaper.bbl} 
\end{document}